%% file: main.tex
\documentclass[conference]{IEEEtran}
\IEEEoverridecommandlockouts
\usepackage{cite}
\usepackage{newclude}
\usepackage{amsmath,amssymb,amsfonts}
\usepackage{algorithmic}
\usepackage{graphicx}
\usepackage{textcomp}
\usepackage{xcolor}
\usepackage{url}
\usepackage{hyperref}
\usepackage{threeparttable}
\usepackage{dblfloatfix}
\def\BibTeX{{\rm B\kern-.05em{\sc i\kern-.025em b}\kern-.08em
    T\kern-.1667em\lower.7ex\hbox{E}\kern-.125emX}}
\begin{document}

\title{Hierarchical Vision Transformers for Cardiac Ejection Fraction Estimation}

\author{\IEEEauthorblockN{
Lhuqita Fazry\textsuperscript{*}, Asep Haryono\textsuperscript{*}, Nuzulul Khairu Nissa\textsuperscript{*}, Sunarno\textsuperscript{*},\\
Naufal Muhammad Hirzi\textsuperscript{*},
Muhammad Febrian Rachmadi\textsuperscript{* +}, Wisnu Jatmiko\textsuperscript{*}
}
\IEEEauthorblockA{
\textsuperscript{*}\textit{Faculty of Computer Science, Universitas Indonesia, Depok, Indonesia} \\
\textsuperscript{+}\textit{RIKEN Research, Tokyo, Japan} \\
lhuqita.fazry@ui.ac.id
}
}

\maketitle
\include*{sections/abstract}
\include*{sections/introduction}
\include*{sections/related_work} 
\include*{sections/method} 
\include*{sections/experiments} 
\include*{sections/result} 
\include*{sections/conclusion}
\include*{sections/acknowledgment}
\bibliographystyle{IEEEtran}
\bibliography{references}
\end{document}

%% file: sections/abstract.tex
\begin{abstract}
The left ventricular of ejection fraction is one of the most important metric of cardiac function. It is used by cardiologist to identify patients who are eligible for life-prolonging therapies. However, the assessment of ejection fraction suffers from inter-observer variability. To overcome this challenge, we propose a deep learning approach, based on hierarchical vision Transformers, to estimate the ejection fraction from echocardiogram videos. The proposed method can estimate ejection fraction without the need for left ventrice segmentation first, make it more efficient than other methods. We evaluated our method on EchoNet-Dynamic dataset resulting $5.59$, $7.59$ and $0.59$ for MAE, RMSE and $\text{R}^2$ respectivelly. This results are better compared to the state-of-the-art method, Ultrasound Video Transformer (UVT). The source code is available on \href{https://github.com/lhfazry/UltraSwin}{https://github.com/lhfazry/UltraSwin}.
\end{abstract}


%% file: sections/introduction.tex
\section{Introduction}
The cardiovascular system is the human circulatory system consists of various important organs which have the main function to circulate oxygen, nutrients, and hormones to all cells and tissues of the body \cite{Tringelova2007}. One of the vital organs in the circulatory system is the cardiac which pump blood throughout the body and receive blood flow back. Based on data from the World Health Organization (WHO), cardiovascular disease is still a deadly disease worldwide. Every year the death rate from this disease increases and in 2019 around 17.9 million people died or 32\% of the world's mortality rate \cite{Organization2021}. Therefore, a fast and accurate method is needed for cardiac diagnoses so it can be handled quickly and properly.

A common method to diagnose cardiac disease is the assesment through echocardiograph video. It is an imaging technique to assess the cardiac function and structure \cite{101007}. The information that is taken from echocardiograph video can be used as the basis for initial screening to diagnoses the cardiac disease. It can also helps for deciding further treatments.

One of the most important metric that can be used to determine the cardiac function is Left Ventricular Ejection Fraction (LVEF) or Ejection Fraction (EF) for short \cite{Wood2014}. EF measures how much blood volume that are ejected out of cardiac within one heart-beat. To calculate EF from echocardiograph video, a cardiologist need to tracing the left ventricular to estimate End Systolic Volume (ESV) and End Diastolic Volume (EDV). ESV is the volume of left ventricular after the ejection process. On the other hand, EDV is the volume of left ventricular before the ejection process. Having the value of ESV and EDV in hand, EF is then calculated using the following formula:
\begin{align}
    EF &= \frac{EDV-ESV}{EDV} \times 100\% \label{eq:ef-formula}
\end{align}

EF can be used to classify the cardiac condition using common threshold. EF value which is less than 50\% can be considered as cardiomyopathies \cite{Ouyang2020}. Cardiomyopathies are a heterogeneous group of heart muscle diseases and an important cause of heart failure (HF) \cite{petar}. Cardiac with EF less than $50\%$ is an indication of heart failure. Heart failure with preserved ejection fraction (HFpEF) has been defined as having signs and symptoms of heart failure with preserved EF and diastolic abnormalities \cite{Lekavich2015}.

However, manually tracing the left ventricular and calculate the EF is very complicated task. It suffer from inter-observer variability. The EF can varies from one heart-beat to another. Furthermore, the American Society of Echocardiography (ASE) and the European Association of Cardiovascular Imaging (EACVI) recommend to observe up to 5 consecutive heart-beats, thus making the approach more complicated \cite{lang_recommendations_2015}. So the method that can estimate EF faster is needed.

With the advance of deep learning, some methods are developed to overcome this problem. Jahren et. al. use the combination of Convolutional Neural Network (CNN) and Recurrent Neural Network (RNN) to predict the location of end-diastole from electrocardiogram data (ECG) \cite{Jahren2020}. Ouyang et. al. use the combination of 3D convolution and atrous convolution to estimate EF \cite{Ouyang2020}. It's clear from formula \ref{eq:ef-formula}, ESV and EDV are needed to calculate EF. The above methods first segment the left ventricular in an echocardiograph video. From the segmentation, they try to detect ESV and EDV and then estimate the volume. 

Recently,  Reynaud et al. proposed UltraSound Video Transformers (UVT) \cite{101007} to estimate EF from echocardiograph video. UVT uses the Transformers, a popular model in Natural Language Processing (NLP), as a features extraction. Before processed by the Transformers, the input video is splitted frame by frame. Each frame is then encoded by ResNet Auto Encoder (ResNetAE) to reduce the dimension into $1,024$ token length. This low dimensional features are then learned by the Transformers to produce another feature maps. The feature maps are then processed by head regressor to produce EF estimation.

However, the process of learning from low dimensional features are not optimal, because most important feature may be loss during the encoding process. In this paper, we propose a novel method to predict EF by directly process the input video using hierarchical vision Transformers. Our method also can directly estimate the EF without the need to segment ESV and EDV and calculate their respective volume.

\begin{figure*}[t]
\centerline{\includegraphics[width=0.8\textwidth]{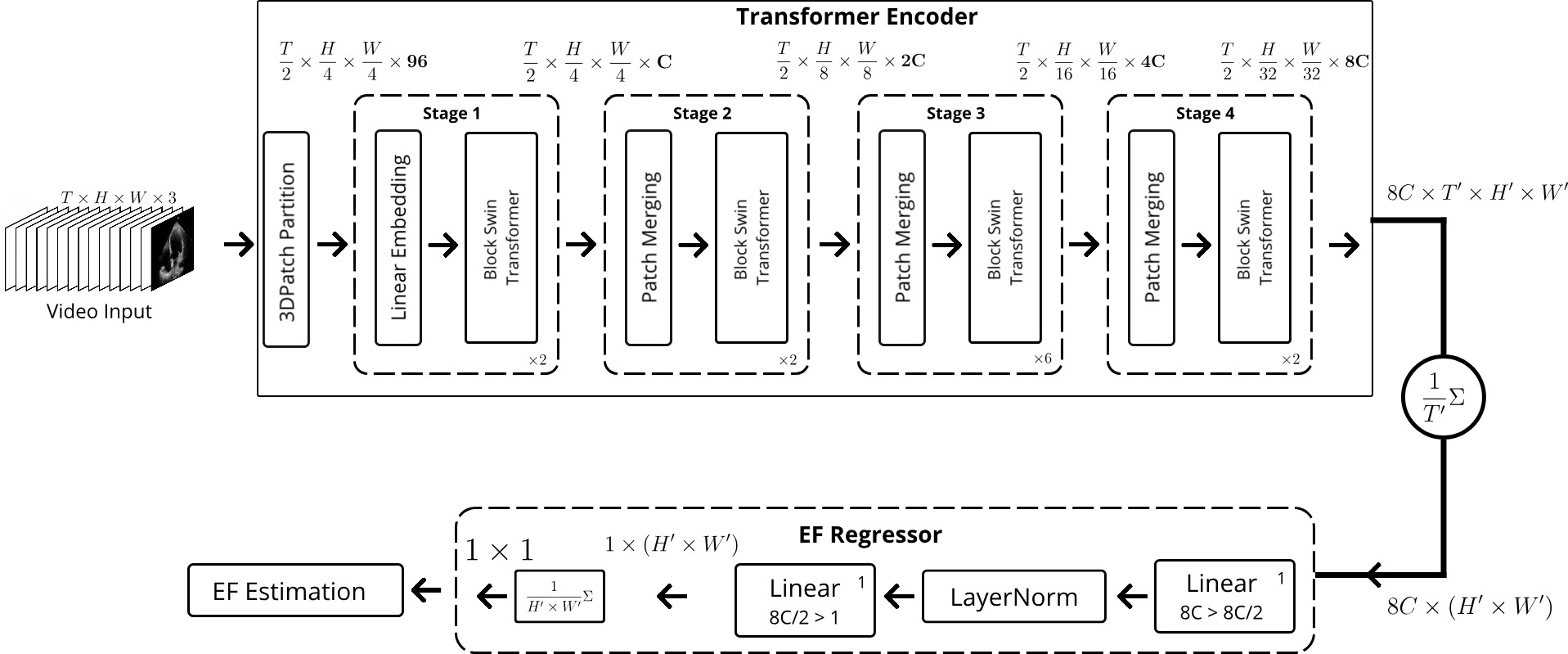}}
\caption{Overall UltraSwin architecture. UltraSwin processes cardiac ultrasound video then output an estimation of ejection fraction for the video. UltraSwin consists of two main modules: Transformers Encoder (TE) and EF Regressor. TE modules acts as a features extractor and EF Regressor as regressor head.}
\label{fig:ultraswin-overall}
\end{figure*}

The focus of this research is to estimate the value of the EF that can be used to diagnose cardiomyopathy (the abnormalities in the heart muscle that can cause heart failure), assess eligibility for certain chemotherapies, and determine indication for medical devices \cite{Ouyang2019}. The output of the regression task that utilize deep learning model in this study is the EF value. The common threshold value on the EF which is less than 50\% can be used to classify cardiomyopathy, this threshold will be used as a reference to determine the heart condition \cite{Ouyang2020}.  

%% file: sections/related_work.tex
\section{Related Work}
Video data is any sequence of time-varying images. In the video data, the picture information is digitized both spatially and temporally. Nowadays, research on video data processing is also an emerging field of computer vision (CV) \cite{video_proc2021}.

Furthermore, video processing techniques have begun to be used in research in the field of medical imaging. One of them is Ghorbani et al. that uses the CNN method with the architecture based on Inception-Resnet-v1 \cite{Ghorbani2020}. Inception-Resnet-v1 has good performance on the Imagenet benchmark dataset and computationally efficient compared to other architectures \cite{Szegedy2016}. This research proved that the use of the deep learning method applied to echocardiography was able to identify the cardiac local anatomy and structure, to estimate metrics for measuring cardiac function and to predict the characteristics of the patients such as gender, height, and weight that not easily observed by human \cite{Ghorbani2020}. 

Other research from \cite{Howard2020} comparing the use of four CNN architectures which aims to classify 14 classes of echocardiographic views consisting of single frame classification (2D CNN), multi-frame classification (TD CNN), spatio-temporal convolution (3D CNN) and two stream classifications. The best-performing model was a "two-stream" network using both spatial and optical flow inputs, with a corresponding error rate $3.9\%$.

Ouyang et al. used the model of spatio-temporal convolutions with residual connections and generates frame-level to predict the EF for each cardiac cycle and then generates frame-level semantic segmentations of the left ventricle using weak supervision from expert human tracings. These outputs are combined to create beat-to-beat predictions of the EF and to predict the presence of heart failure. This study uses echocardiography video dataset from EchoNet-Dynamic \cite{Ouyang2019}. 

In other research \cite{101007} which can also perform the task of predicting EF values by utilizing the Transformers based architecture that capable to process the videos of arbitrary duration. The method uses a Transformers architecture based on the Residual Auto-Encoder (ResAE) Network and a BERT model adapted for token classification.

Based on Reynaud et al \cite{101007}, it can be concluded that the backbone architecture modeling in computer vision (CV) has begun to shift to the use of the Transformers architecture. The trend started with the introduction of  ViT (Vision Transformers), which globally models non-overlapping spatial relationships in image patches using the standard Transformers encoder \cite{Vaswani2017}. For this research, we use Video Swin Transformers \cite{Liu2021}, which completely follows the original Swin Transformers hierarchical structure \cite{Liu2022}. However, we extend the local attention computation scope from the spatial domain to the spatio-temporal domain. The adaptation process was carried out in the 3D patch partition section and replaced the local window self-attention module into a 3D shifted window based on multi-head self-attention (MSA) and shifted window multi-head self-attention (SW-MSA) in the Transformers Block section. Video Swin Transformers can do the video-recognition tasks that contains an inductive bias towards spatio-temporal locality. 

%% file: sections/method.tex
\section{Method}

\begin{figure*}[t]
\centerline{\includegraphics[width=0.6\textwidth]{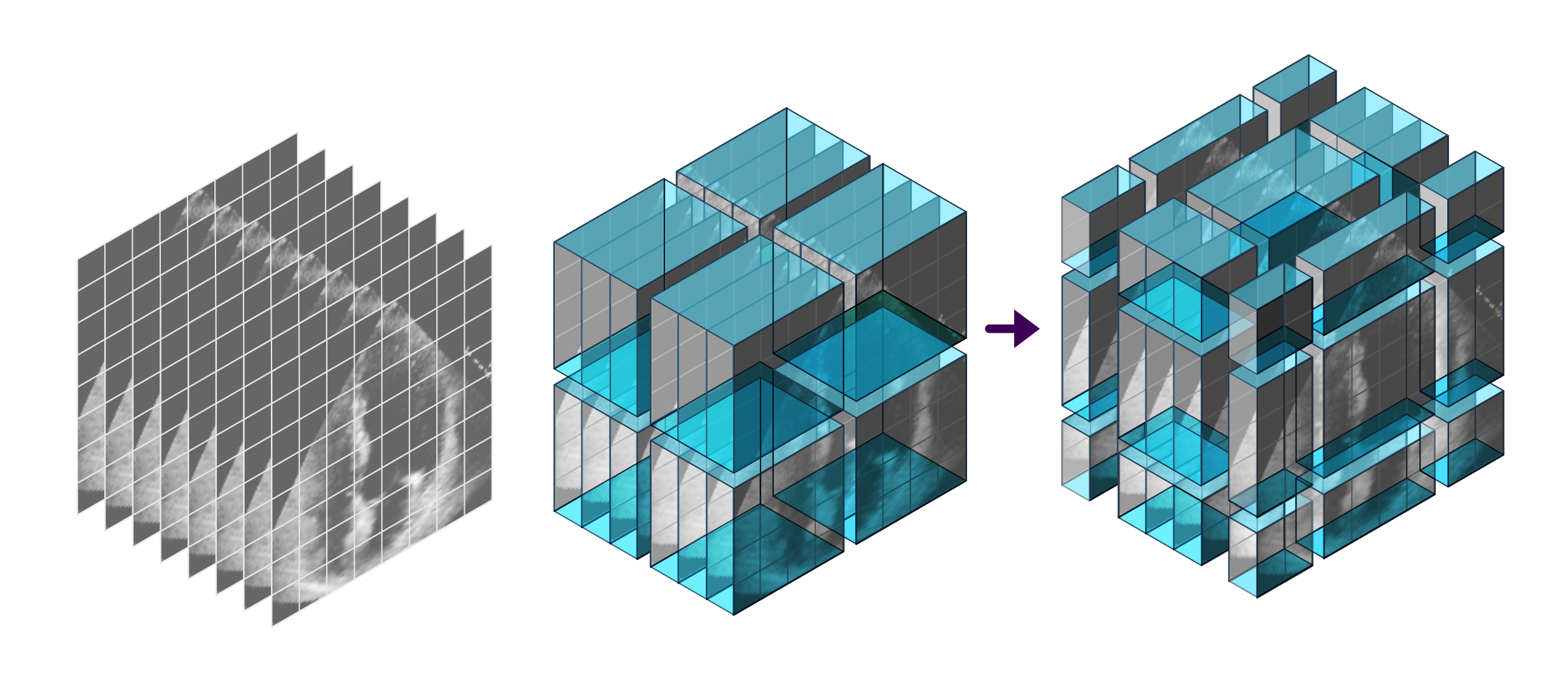}}
\caption{The illustration of 3D tokens and shifted windows mechanism. At first, each frame is splitted into patches. The patches are then grouped into windows. In the two consecutive attention layers, the windows configuration are then shifted. In this way, the attention can happen across windows while keeping the computation cheap, because the attention are only calculated within window (not calculated globally)}.
\label{fig:3dtoken}
\end{figure*}

In this paper,  we propose a novel method to estimate EF from a cardiac ultrasound video. Our method uses a deep learning model, based on hierarchical vision Transformers. We named the model as UltraSwin. UltraSwin adopts Transformers \cite{Vaswani2017}, a popular deep learning model in Natural Language Processing (NLP) and its derivative work on Computer Vision (CV) \cite{Dosovitskiy2020, Liu2022}.

\subsection{Model Architecture}
The architecture of UltraSwin is described in figure \ref{fig:ultraswin-overall}. The model is received a video as an input, specifically ultrasound video containing short time cardiac recording. The output of the model is the estimation of EF for the cardiac in the ultrasound video.

UltraSwin has 2 main modules, Transformers Encoder (TE) and EF Regressor. The TE module acts as feature extractor while EF Regressor as regressor head. The TE module is used to learn representation from input video and then output the feature maps. They are then processed by EF Regressor and transformed into scalar value. This value is then used as an EF estimation for the input ultrasound video.

Instead of treat a frame of the video as input token like in UVT \cite{101007}, UltraSwin uses 3D video patches as input token following work of Liu et. al.\cite{Liu2021}.

\subsection{Pre-processing}
In this research, we use ultrasound video from EchoNet-Dynamic dataset \cite{Ouyang2019}. This dataset contains echocardiography videos with variety of frame length and contains at least one heart-beat. Although, it can have more than one heart-beat per video, ES (End Systolic) and ED (End Dyastolic) ground-truth are only for one heart-beat. 

Each frame in the video have spatial dimension of $112 \times 112$ pixels. The frame's width and height must satisfy $2^n$, so it can be processed into patches. Each input for Transformers must have same length, so we cut the video into fixed length of $128$ frames. We choose $128$, because it is the closest $2^n$ value from $112$. We select ES and ED frames and any frames between them then cut them. The Echonet Dynamic dataset that we used in this research, contains varied length (total frames), frame rate and image quality \cite{Ouyang2019}. Therefore, if the total frames in the cut out video are more than $128$, we subsample the frames between ES and ED. Otherwise, we repeat or mirroring the frames between ES and ED and place them after ED to get the total $128$ frame length. Suppose the sequence of frames $F = [m_{ES}, m_{b_1},\cdots{}, m_{b_n}, m_{ED}]$, we repeat the frames between ED and ES to create new sequences $\hat{F} = [m_{ES}, m_{b_1},\cdots{}, m_{b_n}, m_{ED}, m_{b_1},\cdots{}]$. We choose this technique based on the research by Reynaud et al., where the mirroring technique gives better results than the random sampling technique in terms of fitting the number of total frames to 128 \cite{101007}. After that, we pad the frames with blank pixels, so its dimension becomes $128 \times 128$ pixels.

We also tried to augment the video with standard augmentations like horizontal flip, vertical flip, random rotation and others. Suprisingly, we found that augmentations lead to worse performance. This result indicates that ultrasound video dataset are sensitive to augmentation operations.

\subsection{Transformers Encoder}
This module contains 4 stages. Unlike Vision Transformers (ViT) \cite{Dosovitskiy2020} that has fixed patch size along the stages, UltraSwin use hierarchical architecture following Swin Transformers \cite{Liu2022} in the spatial spaces. At every stage, the patch size is downsampled into half of the patch size in the previous stages. To make the model learn temporal information, UltraSwin follows Video Swin Transformers \cite{Liu2021} to process the video input in the shape the 3D patches. 

TE module contains two main components.

\subsubsection{3D Patch Partition}
Suppose an input video has dimension of $T \times H \times W \times 3$, where $T, H, W$ and $3$ represent number of frames, frame's height, frame's width and number of channels respectively. The video input is then partitioned into 3D patch with dimension $2 \times 4 \times 4 \times 3$. In the Transformers world, this 3D patch is called token. Each token contains embedding features with length $96$. Actually, we can use any number other than $96$, but greater number can significantly affects the computation cost. We use $96$ following Video Swin Transformers\cite{Li2021} as it gives a good performance. This process yields $\frac{T}{2} \times \frac{H}{4} \times \frac{W}{4}$ tokens in total. The tokens are then flattened into sequences before processed by the Transformers. Figure \ref{fig:3dtoken} illustrates the 3D tokens.

The features of each token are then transformed by a linear layer into an arbritrary $C$ dimensions. So, the dimension of the tokens is now $\frac{T}{2} \times \frac{H}{4} \times \frac{W}{4} \times C$. This number is hyper-parameter and we can use arbitrary number for $C$.

\subsubsection{Block Swin Transformers}
Transformers \cite{Vaswani2017} and Vision Transformers (ViT) \cite{Dosovitskiy2020} use global self-attention (SA) and compute softmax score between each tokens, thus making the computation and memory resources grow quadratically with token length. This approach is efficient enough for single input image. On the other side, the video have multiple image frames, so the approach are not suitable for video related tasks like video classification, video segmentation and others. UltraSwin use local window self-attention following Swin Transformers \cite{Liu2022} that is proven more efficient in video related task than global self-attention \cite{Li2021}.

While efficient, local window self-attention lacks of connection accross window. This can cause performance degradation on the model. To solve this issue, UltraSwin shifts the window partition in two consecutive Swin Transformers Block as illustrated in figure \ref{fig:3dtoken}. As Transformers can have multiple layers of blocks, UltraSwin shifted the window configuration in every two consecutive blocks. This design is proven to be effective in image recognition task \cite{Liu2022}. The main reason why it is effective is because it enables the connections between non-overlapping windows with their neighbours.

Suppose a sequence of 3D tokens with size $T^{'} \times H^{'} \times W^{'} \times 3$. In the first layer, these tokens are then arranged into regular non-overlapping window of size $P \times M \times M$, thus resulting $\lceil{}\frac{T^{'}}{P}\rceil{} \times \lceil{}\frac{H^{'}}{M}\rceil{} \times \lceil{}\frac{W^{'}}{M}\rceil{}$ non-overlapping 3D windows in total. In the second layer, configuration of every window is shifted within width, height and temporal axes by $(\frac{P}{2} \times \frac{M}{2} \times \frac{M}{2})$.

The self-attention mechanism is applied multiple times in parallel. This is called heads. In multi-head scenario, the output from each self-attention are concatenated. In first layer, we called multi-head self-attention (MSA) and shifted window multi-head self-attention (SW-MSA) for second layer. Formally, we stated MSA as $[\text{SA}_1, \text{SA}_2, \cdots{}, \text{SA}_n]$ and SW-MSA as $[\text{SW-SA}_1, \text{SW-SA}_2, \cdots{}, \text{SW-SA}_n]$, where $\text{SA}_i$ and $\text{SW-SA}_i$ refer to self-attention in layer-$i$ and shifted-window self-attention in layer-$i$ respectivelly. The SA itself can be formulated as follow:
\begin{align}
    \text{SA}(Q,K,V) &= \text{SoftMax}(\frac{QK^T}{\sqrt{d}} + B)V
\end{align}
where $K,V,Q \in \mathbb{R}^{PM^2 \times d}$ are matrices for \emph{key}, \emph{value} and \emph{query} respectively, while $d$ is \emph{query} and \emph{key} dimension, $PM^2$ is the number of tokens in 3D window, and $B \in \mathbb{R}^{P^2 \times M^2 \times M^2}$ is matrix of relative position bias.

\begin{table}[ht]
    \centering
    \begin{threeparttable}
    \caption{UltraSwin variants}
    \label{tab:model-variant}
    \begin{tabular}{|l|c|c|}
         \hline 
         \textbf{Parameter} & \textbf{UltraSwin-base} &  \textbf{UltraSwin-small}\\
         \hline
         embedding dimension & 128 & 96\\
         \hline
         number of head & 4, 8, 16, 32\tnote{1}& 3, 6, 12, 24\tnote{1}\\
         \hline
         layer depth & 2, 2, 18, 2\tnote{1} & 2, 2, 18, 2\tnote{1}\\
         \hline
         Total parameter & 88.2M & 49.7M\\
         \hline
    \end{tabular}
    \begin{tablenotes}
    \item[1] These values are for stage 1, 2, 3 and 4 respectivelly
    \end{tablenotes}
    \end{threeparttable}
\end{table}

The self-attention blocks are then followed by feed forward networks, which is 2 layers MLP with GELU \cite{Hendrycks2016} non-linearity in between. Layer Normalization (LN) \cite{Ba2016} is applied before self-attention module and before the MLP. Residual connection \cite{7780459} is then applied after self-attention and after MLP. Two consecutives of Swin Transformers blocks in layer-$l$ and layer-$l+1$ can be formulated as follow:
\begin{align}
    \hat{z}^l &= \text{MSA}(\text{LN}(z^{l-1})) + z^{l-1}  \nonumber \\
    z^l &= \text{MLP}(\text{GELU}(\text{LN}(\hat{z}^l))) + \hat{z}^l  \nonumber \\
    \hat{z}^{l+1} &= \text{SW-MSA}(\text{LN}(z^{l})) + z^{l}  \nonumber \\
    z^{l+1} &= \text{MLP}(\text{GELU}(\text{LN}(\hat{z}^{l+1}))) + \hat{z}^{l+1}
\end{align}

\subsection{EF Regressor}
The EF regressor take the output of the TE module as input. The input is a features map with dimension $\frac{T}{2} \times \frac{H}{32} \times \frac{W}{32} \times 8C$. The temporal axes are then reduced from the map, resulting new dimension $\frac{H}{32} \times \frac{W}{32} \times 8C$. A linier layer is then applied to reduce the last axes of feature maps from $8C$ into $4C$. A Layer Normalization (LN) is then applied followed by linier layer to reduce the feature axes into $1$ dimension. Spatial reduction is then applied to the map resulting $1 \times 1$ dimesion scalar. This scalar value is then used as EF estimation.

\subsection{Model Variants}
We propose two variants of UltraSwin: UltraSwin-base and UltraSwin-small. Table \ref{tab:model-variant} summarizes the two variants. Number of head and layer depth values in table \ref{tab:model-variant} refer to configurations on stage $1, 2, 3$ and $4$ respectivelly. The total parameter for UltraSwin-small is almost half from total parameter for UltraSwin-base.

\subsection{Loss Function}
Both variants are trained to minimize MSE (Mean Squared Error). We use MSE because it is commonly used in regression task and gives the best performances. MSE is defined as follow:
\begin{align}
    L(y,\hat{y}) &= \frac{1}{N} \sum_{i=1}^{N}\left(y_i - \hat{y}_i\right)^2
\end{align}
where $y_i$ and $\hat{y}_i$ refer to EF ground-truth and EF prediction from the model respectivelly.

%% file: sections/experiments.tex
\section{Experiments}
\subsection{Dataset}
The dataset used in this research is EchoNet-Dynamic \cite{Ouyang2019} which is an open dataset and sourced from the Stanford Artificial Intelligence in Medicine and Imaging (AIMI) Center obtained from \href{https://stanfordaimi.azurewebsites.net}{https://stanfordaimi.azurewebsites.net}. The dataset contains videos of heart movement and chamber volume from echocardiography or cardiac ultrasound. The total video is $10,030$ in .avi format consists of $7,465$ training data, $1,288$ validation data and $1,277$ test data. 

Each video has varies duration with number of frames ranging from $28$ to $1002$. The spatial dimension for each frame is $112 \times 112$ pixels. Each video has frame per second (FPS) $50$. The videos come with EF ground truth which value ranging from $6.9$ to $96.96$.

\subsection{Implementation Details}
The model architecture was created using the \texttt{Python 3.8} programming language and the \texttt{PyTorch 1.11} framework. The \texttt{Pytorch Lightening 1.6.4} library was used to simplify the training process. We also use Tensorboard library to records the evaluation metrics. The model was trained using a $1$ core NVidia Tesla T4 GPU. To save the memory usage, $16$ bit precision is used for gradient calculations during training and \texttt{batch\_accumulation} $=2$ to speed up the training process.

In the UltraSwin-base model, the \texttt{batch\_size} parameter used is $2$, while in the UltraSwin-small is $4$. To speed up the model convergencies during training, we initialize the TE module weights using the pre-trained Swin Transformer model that had been trained using the ImageNet 22k dataset \cite{Deng2010}.

\subsection{Training Details}
The UltraSwin models were trained without freezing the TE module to avoid the problem of different domains in transfer learning. For the UltraSwin-base, the TE module weights are initialized using pre\-trained \texttt{swin\_base\_patch4\_ window7\_224\_22k}, while for the UltaSwin-small using the pre\-trained \texttt{swin\_small\_patch4\_window7\_224\_22k}. Both pre\-trained models can be downloaded at the \url{https://github.com/microsoft/Swin-Transformer} page.

During the training process, AdamW \cite{AdamW} optimization was used with an initial learning rate of $10^{-4}$ and weight decay of $10^{-4}$. Both models were trained for $20$ epochs. At each epoch, the learning rate was reduced by $0.15$ from the learning rate in the previous epoch. 

On the UltraSwin-base model, the training process takes approximately $30$ minutes for one epoch, while on the UltraSwin-small it takes approximately $15$ minutes for one epoch. When making predictions using the trained model, the same configuration is used as the configuration in the training. However, because the inference process only performs forward propagation without the need to calculate the gradient (back propagation), the \texttt{batch\_size} parameter can be increased to $8$.

%% file: sections/result.tex
\section{Result and Discussion}
Here we show the result of the experiments for UltraSwin-base and UltraSwin-small. We then compared the results of our two variations models with the state-of-the-art method, Ultrasound Video Transformers (UVT) \cite{101007}.

\begin{table}[ht]
    \centering
    \caption{Result comparison of UltraSwin and UVT}.
    \label{tab:hasil}
    \begin{tabular}{|l|c|c|c|c|}
         \hline 
         \textbf{Model} & \textbf{Total Parameter} & \textbf{MAE} &  \textbf{RMSE} &  $\mathbf{R^2}$\\
         \hline
         UVT & $346.8$M & $5.95$ & $8.38$ & $0.52$\\
         \hline
         UltraSwin-small  &$49.7$M & $5.72$ & $7.63$ & $0.58$\\
         \hline
         UltraSwin-base & $88.2$M & $\mathbf{5.59}$ & $\mathbf{7.59}$ & $\mathbf{0.59}$ \\
         \hline
    \end{tabular}
\end{table}

Table \ref{tab:hasil} summarizes the result of our experiments and we compare it with the results of UVT model from Reynaud et al \cite{101007}. We use three metrics to evaluate the models, MAE (Mean Absolute Error), RMSE (Root Mean Squared Error) and $R^2$ (Coefficient of Determination). Smaller value of MAE and RMSE means better performance. However, higher value of $R^2$ means better performance. 

It can be seen that UltraSwin-small with smaller number of parameters than UVT is able to produce a smaller values for MAE and RMSE and higher value for $R^2$. This proves that UltraSwin-small is superior to UVT. Furthermore, UltraSwin-base is superior to UltraSwin-small. Both variations of UltraSwin are able to outperform the UVT on the three evaluation metrics.

\begin{figure}[h]
    \centering
    \includegraphics[width=5cm]{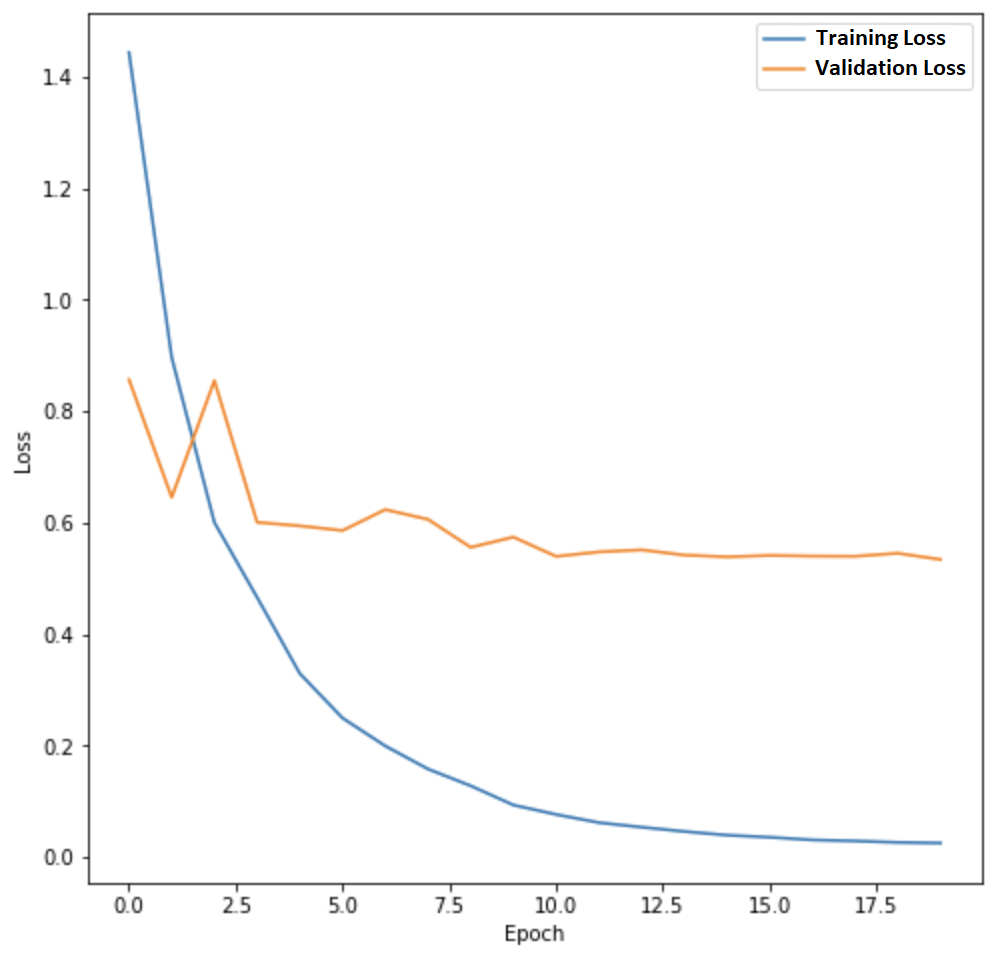}
    \caption{Graph of training and validation loss for UltraSwin-small. It can be seen from this graph that both training and validation loss are reduced as epoch increases. On the early epoch, the validation loss is fluctuated. It is because the model still in early learning phase. After 3 epochs, the validation is quite stable.}
    \label{fig:ultraSwin-small}
\end{figure}

During training, we log the training and validation losses at every epoch. Figure \ref{fig:ultraSwin-small} shows the training and validation losses for UltraSwin-small model. The blue and orange line represent training and validation losses. From the graph, it can be seen that both training and validation losses are reduced as the epoch increased. But the validation loss seems to be fluctuated on early epoch for UltraSwin-small. It's because the model still in the early learning phase. After 3 epochs, the reduction of validation loss is quite stable.

\begin{figure}[h]
    \centering
    \includegraphics[width=5cm]{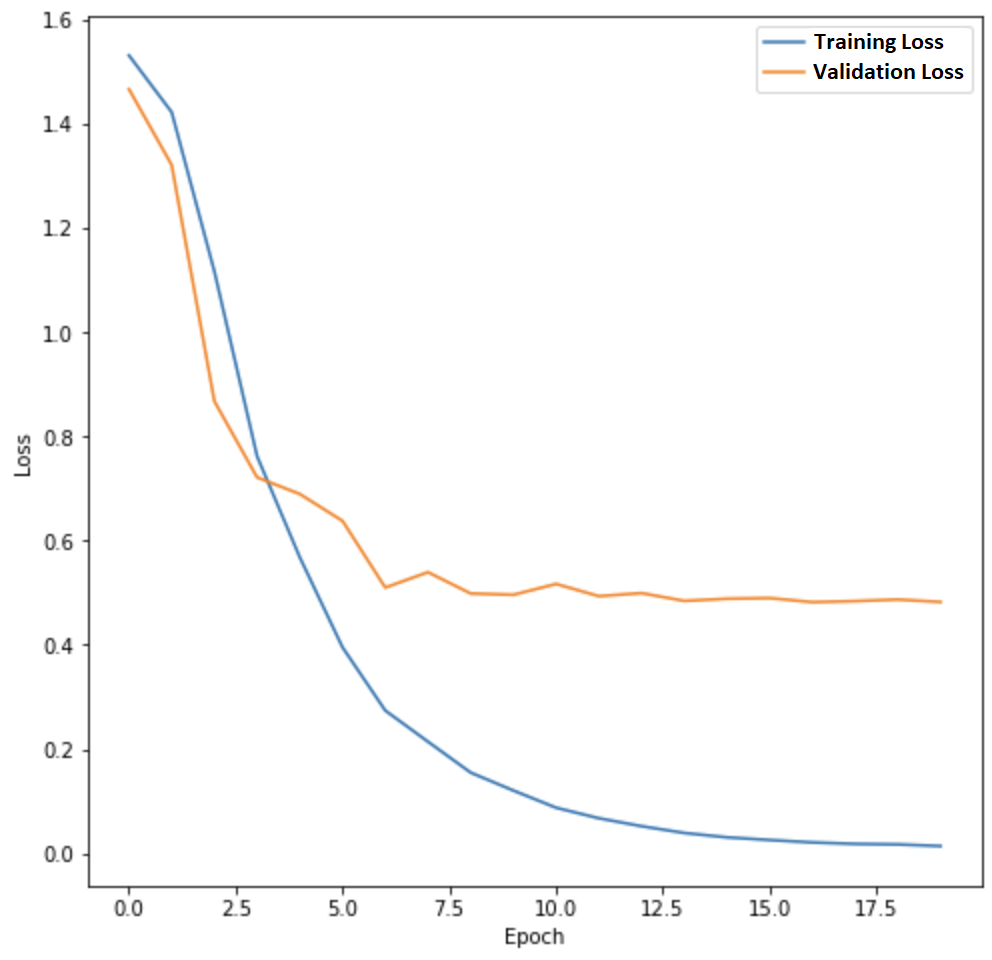}
    \caption{Graph of training and validation loss for UltraSwin-base. It can be seen from this graph that both training and validation loss are reduced as epoch increases. It indicates that the model learn very well}
    \label{fig:ultraSwin-base}
\end{figure}

Similar to UltraSwin-small, both training and validation losses for UltraSwin-base are reduced as the epoch increased. It can be seen from Figure \ref{fig:ultraSwin-base} that the loss reduction are quite stable both for training and validation loss.

%% file: sections/conclusion.tex
\section{Conclusion}
In this paper, we propose UltraSwin, a novel method to estimate EF from echocardiogram videos. This method uses Swin Transformers, a hierarchical vision transformers to extract spatio-temporal features. Furthermore, it gives better EF estimation than UVT. One can futher research to improve UltraSwin performance, for example by aggregating the features extraction on every stages before processed by EF regressor or use combination between 3D tokens and another Vision Transformers backbone such as Pyramid Vision Transformer (PVTv2) \cite{pvt2}.

%% file: sections/acknowledgment.tex
\section*{Acknowledgment}
This work is supported by Research Laboratory of Faculty of Computer Science, Universitas Indonesia. Thank you for contributing to provide some facilities in laboratory and supporting this research.

%% file: main.bbl
\begin{thebibliography}{10}
\providecommand{\url}[1]{#1}
\csname url@samestyle\endcsname
\providecommand{\newblock}{\relax}
\providecommand{\bibinfo}[2]{#2}
\providecommand{\BIBentrySTDinterwordspacing}{\spaceskip=0pt\relax}
\providecommand{\BIBentryALTinterwordstretchfactor}{4}
\providecommand{\BIBentryALTinterwordspacing}{\spaceskip=\fontdimen2\font plus
\BIBentryALTinterwordstretchfactor\fontdimen3\font minus
  \fontdimen4\font\relax}
\providecommand{\BIBforeignlanguage}[2]{{%
\expandafter\ifx\csname l@#1\endcsname\relax
\typeout{** WARNING: IEEEtran.bst: No hyphenation pattern has been}%
\typeout{** loaded for the language `#1'. Using the pattern for}%
\typeout{** the default language instead.}%
\else
\language=\csname l@#1\endcsname
\fi
#2}}
\providecommand{\BIBdecl}{\relax}
\BIBdecl

\bibitem{Tringelova2007}
M.~Tringelov{\'{a}}, P.~Nardinocchi, L.~Teresi, and A.~{Di Carlo}, ``{The
  cardiovascular system as a smart system},'' \emph{Topics on Mathematics for
  Smart Systems - Proceedings of the European Conference}, no. January, pp.
  253--270, 2007.

\bibitem{Organization2021}
\BIBentryALTinterwordspacing
W.~H. Organization~(WHO), ``{Cardiovascular diseases (CVDs)},'' 2021. [Online].
  Available:
  \url{https://www.who.int/en/news-room/fact-sheets/detail/cardiovascular-diseases-(cvds)}
\BIBentrySTDinterwordspacing

\bibitem{101007}
H.~Reynaud, A.~Vlontzos, B.~Hou, A.~Beqiri, P.~Leeson, and B.~Kainz,
  ``Ultrasound video transformers for cardiac ejection fraction estimation,''
  in \emph{Medical Image Computing and Computer Assisted Intervention -- MICCAI
  2021}, M.~de~Bruijne, P.~C. Cattin, S.~Cotin, N.~Padoy, S.~Speidel, Y.~Zheng,
  and C.~Essert, Eds.\hskip 1em plus 0.5em minus 0.4em\relax Cham: Springer
  International Publishing, 2021, pp. 495--505.

\bibitem{Wood2014}
P.~W. Wood, J.~B. Choy, N.~C. Nanda, and H.~Becher, ``{Left ventricular
  ejection fraction and volumes: it depends on the imaging method.}''
  \emph{Echocardiography}, vol.~70, pp. 87--100, 2014.

\bibitem{Ouyang2020}
\BIBentryALTinterwordspacing
D.~Ouyang, B.~He, A.~Ghorbani, N.~Yuan, J.~Ebinger, C.~P. Langlotz, P.~A.
  Heidenreich, R.~A. Harrington, D.~H. Liang, E.~A. Ashley, and J.~Y. Zou,
  ``Video-based ai for beat-to-beat assessment of cardiac function,''
  \emph{Nature}, vol. 580, pp. 252--256, 2020. [Online]. Available:
  \url{http://dx.doi.org/10.1038/s41586-020-2145-8}
\BIBentrySTDinterwordspacing

\bibitem{petar}
P.~Seferovic, M.~Polovina, J.~Bauersachs, M.~Arad, T.~{Ben Gal}, L.~Lund,
  S.~Felix, E.~Arbustini, A.~Caforio, D.~Farmakis, G.~Filippatos, E.~Gialafos,
  V.~Kanjuh, G.~Krljanac, G.~Limongelli, A.~Linhart, A.~Lyon, R.~Maksimovic,
  D.~Milicic, I.~Milinkovic, M.~Noutsias, A.~Oto, O.~Oto, S.~Pavlovic,
  M.~Piepoli, A.~Ristic, G.~Rosano, H.~Seggewiss, M.~Asanin, J.~Seferovic,
  F.~Ruschitzka, J.~Celutkiene, T.~Jaarsma, C.~Mueller, B.~Moura, L.~Hill,
  M.~Volterrani, Y.~Lopatin, M.~Metra, J.~Backs, W.~Mullens, O.~Chioncel,
  R.~{de Boer}, S.~Anker, C.~Rapezzi, A.~Coats, and C.~Tschoepes,
  ``\BIBforeignlanguage{English}{Heart failure in cardiomyopathies: a position
  paper from the heart failure association of the european society of
  cardiology},'' \emph{\BIBforeignlanguage{English}{European Journal of Heart
  Failure}}, vol.~21, no.~5, pp. 553--576, May 2019.

\bibitem{Lekavich2015}
\BIBentryALTinterwordspacing
C.~L. Lekavich, D.~J. Barksdale, V.~Neelon, and J.~R. Wu, ``{Heart failure
  preserved ejection fraction (HFpEF): an integrated and strategic review},''
  \emph{Heart Failure Reviews}, vol.~20, no.~6, pp. 643--653, nov 2015.
  [Online]. Available:
  \url{https://link.springer.com/article/10.1007/s10741-015-9506-7}
\BIBentrySTDinterwordspacing

\bibitem{lang_recommendations_2015}
\BIBentryALTinterwordspacing
R.~M. Lang, L.~P. Badano, V.~Mor-Avi, J.~Afilalo, A.~Armstrong, L.~Ernande,
  F.~A. Flachskampf, E.~Foster, S.~A. Goldstein, T.~Kuznetsova, P.~Lancellotti,
  D.~Muraru, M.~H. Picard, E.~R. Rietzschel, L.~Rudski, K.~T. Spencer,
  W.~Tsang, and J.-U. Voigt, ``\BIBforeignlanguage{en}{Recommendations for
  {Cardiac} {Chamber} {Quantification} by {Echocardiography} in {Adults}: {An}
  {Update} from the {American} {Society} of {Echocardiography} and the
  {European} {Association} of {Cardiovascular} {Imaging}},''
  \emph{\BIBforeignlanguage{en}{Journal of the American Society of
  Echocardiography}}, vol.~28, no.~1, pp. 1--39.e14, Jan. 2015. [Online].
  Available:
  \url{https://linkinghub.elsevier.com/retrieve/pii/S0894731714007457}
\BIBentrySTDinterwordspacing

\bibitem{Jahren2020}
T.~S. Jahren, E.~N. Steen, S.~A. Aase, and A.~H. Solberg, ``Estimation of
  end-diastole in cardiac spectral doppler using deep learning,'' \emph{IEEE
  Transactions on Ultrasonics, Ferroelectrics, and Frequency Control}, vol.~67,
  pp. 2605--2614, 2020.

\bibitem{Ouyang2019}
D.~Ouyang, B.~He, A.~Ghorbani, M.~P. Lungren, E.~A. Ashley, D.~H. Liang, and
  J.~Y. Zou, ``Echonet-dynamic: a large new cardiac motion video data resource
  for medical machine learning,'' \emph{33rd Conference on Neural Information
  Processing Systems (NeurIPS 2019)}, pp. 1--11, 2019.

\bibitem{video_proc2021}
V.~Sharma, M.~Gupta, A.~Kumar, and D.~Mishra, ``Video processing using deep
  learning techniques: A systematic literature review,'' \emph{IEEE Access},
  vol.~9, pp. 139\,489--139\,507, 2021.

\bibitem{Ghorbani2020}
\BIBentryALTinterwordspacing
A.~Ghorbani, D.~Ouyang, A.~Abid, B.~He, J.~H. Chen, R.~A. Harrington, D.~H.
  Liang, E.~A. Ashley, and J.~Y. Zou, ``Deep learning interpretation of
  echocardiograms,'' \emph{npj Digital Medicine}, vol.~3, pp. 1--10, 2020.
  [Online]. Available: \url{http://dx.doi.org/10.1038/s41746-019-0216-8}
\BIBentrySTDinterwordspacing

\bibitem{Szegedy2016}
\BIBentryALTinterwordspacing
C.~Szegedy, V.~Vanhoucke, S.~Ioffe, J.~Shlens, and Z.~Wojna, ``Rethinking the
  inception architecture for computer vision,'' in \emph{2016 IEEE Conference
  on Computer Vision and Pattern Recognition (CVPR)}.\hskip 1em plus 0.5em
  minus 0.4em\relax Los Alamitos, CA, USA: IEEE Computer Society, jun 2016, pp.
  2818--2826. [Online]. Available:
  \url{https://doi.ieeecomputersociety.org/10.1109/CVPR.2016.308}
\BIBentrySTDinterwordspacing

\bibitem{Howard2020}
J.~P. Howard, J.~Tan, M.~J. Shun-Shin, D.~Mahdi, A.~N. Nowbar, A.~D. Arnold,
  Y.~Ahmad, P.~McCartney, M.~Zolgharni, N.~W. Linton, N.~Sutaria, B.~Rana,
  J.~Mayet, D.~Rueckert, G.~D. Cole, and D.~P. Francis, ``Improving ultrasound
  video classification: An evaluation of novel deep learning methods in
  echocardiography,'' \emph{Journal of Medical Artificial Intelligence},
  vol.~3, 2020.

\bibitem{Vaswani2017}
\BIBentryALTinterwordspacing
A.~Vaswani, N.~Shazeer, N.~Parmar, J.~Uszkoreit, L.~Jones, A.~N. Gomez,
  L.~Kaiser, and I.~Polosukhin, ``Attention is all you need,'' 6 2017.
  [Online]. Available: \url{http://arxiv.org/abs/1706.03762}
\BIBentrySTDinterwordspacing

\bibitem{Liu2021}
\BIBentryALTinterwordspacing
Z.~Liu, J.~Ning, Y.~Cao, Y.~Wei, Z.~Zhang, S.~Lin, and H.~Hu, ``Video swin
  transformer,'' pp. 1--12, 2021. [Online]. Available:
  \url{http://arxiv.org/abs/2106.13230}
\BIBentrySTDinterwordspacing

\bibitem{Liu2022}
Z.~Liu, Y.~Lin, Y.~Cao, H.~Hu, Y.~Wei, Z.~Zhang, S.~Lin, and B.~Guo, ``Swin
  transformer: Hierarchical vision transformer using shifted windows,'' pp.
  9992--10\,002, 2022.

\bibitem{Dosovitskiy2020}
\BIBentryALTinterwordspacing
A.~Dosovitskiy, L.~Beyer, A.~Kolesnikov, D.~Weissenborn, X.~Zhai,
  T.~Unterthiner, M.~Dehghani, M.~Minderer, G.~Heigold, S.~Gelly, J.~Uszkoreit,
  and N.~Houlsby, ``An image is worth 16x16 words: Transformers for image
  recognition at scale,'' 10 2020. [Online]. Available:
  \url{http://arxiv.org/abs/2010.11929}
\BIBentrySTDinterwordspacing

\bibitem{Li2021}
\BIBentryALTinterwordspacing
Y.~Li, K.~Zhang, J.~Cao, R.~Timofte, and L.~V. Gool, ``Localvit: Bringing
  locality to vision transformers,'' 4 2021. [Online]. Available:
  \url{http://arxiv.org/abs/2104.05707}
\BIBentrySTDinterwordspacing

\bibitem{Hendrycks2016}
\BIBentryALTinterwordspacing
D.~Hendrycks and K.~Gimpel, ``Gaussian error linear units (gelus),'' 6 2016.
  [Online]. Available: \url{http://arxiv.org/abs/1606.08415}
\BIBentrySTDinterwordspacing

\bibitem{Ba2016}
\BIBentryALTinterwordspacing
J.~L. Ba, J.~R. Kiros, and G.~E. Hinton, ``Layer normalization,'' 2016.
  [Online]. Available: \url{http://arxiv.org/abs/1607.06450}
\BIBentrySTDinterwordspacing

\bibitem{7780459}
K.~He, X.~Zhang, S.~Ren, and J.~Sun, ``Deep residual learning for image
  recognition,'' in \emph{2016 IEEE Conference on Computer Vision and Pattern
  Recognition (CVPR)}, 2016, pp. 770--778.

\bibitem{Deng2010}
J.~Deng, W.~Dong, R.~Socher, L.-J. Li, K.~Li, and L.~Fei-Fei, ``Imagenet: A
  large-scale hierarchical image database.''\hskip 1em plus 0.5em minus
  0.4em\relax Institute of Electrical and Electronics Engineers (IEEE), 3 2010,
  pp. 248--255.

\bibitem{AdamW}
\BIBentryALTinterwordspacing
I.~Loshchilov and F.~Hutter, ``Decoupled weight decay regularization,'' 2017.
  [Online]. Available: \url{https://arxiv.org/abs/1711.05101}
\BIBentrySTDinterwordspacing

\bibitem{pvt2}
\BIBentryALTinterwordspacing
W.~Wang, E.~Xie, X.~Li, D.~Fan, K.~Song, D.~Liang, T.~Lu, P.~Luo, and L.~Shao,
  ``Pvtv2: Improved baselines with pyramid vision transformer,'' \emph{CoRR},
  vol. abs/2106.13797, 2021. [Online]. Available:
  \url{https://arxiv.org/abs/2106.13797}
\BIBentrySTDinterwordspacing

\end{thebibliography}
